\documentclass[letterpaper, 10 pt, conference]{ieeeconf}  
\IEEEoverridecommandlockouts  

\usepackage{cite}
\usepackage{amsmath,amssymb,amsfonts}
\usepackage{algorithmic}
\usepackage[graphicx]{realboxes}
\usepackage{textcomp}
\usepackage[table,xcdraw]{xcolor}
\usepackage{multirow}
\usepackage{array}
\usepackage{tikz}
\usepackage{siunitx}
\usepackage{lscape}
\usepackage{adjustbox}
\usepackage{tabularx}
\usepackage{mleftright}
\usepackage{etoolbox}
\usepackage{xcolor}
\usepackage{booktabs}

\usepackage{makecell}
\usepackage{tabto}
\usepackage{longtable}

\usepackage{kotex}

\usepackage{verbatim}
\usepackage{authblk}
\usepackage{import}
\usepackage{color,colortbl}
\usepackage{float}
\usepackage{lipsum,setspace}
\usepackage{lscape}

\def\BibTeX{{\rm B\kern-.05em{\sc i\kern-.025em b}\kern-.08em T\kern-.1667em\lower.7ex\hbox{E}\kern-.125emX}}

\newcommand{\rom}[1]{\uppercase\expandafter{\romannumeral #1\relax}}

\title{\LARGE \bf ACNet: Mask-Aware Attention with Dynamic Context Enhancement for Robust Acne Detection}

\begin{document}

\author{Kyungseo Min\textsuperscript{\rm 1}, Gun-Hee Lee\textsuperscript{\rm 2}, Seong-Whan Lee\textsuperscript{\rm 3}


\thanks{*This work was supported by the Institute for Information \& Communications Technology Promotion (IITP) grant funded by the Government of South Korea (No. 2017-0-00451, Development of BCI-based Brain and Cognitive Computing Technology for Recognizing User’s Intentions using Deep Learning; No. 2019-0-00079, Artificial Intelligence Graduate School Program, Korea University). We thank Samsung Research for generously supporting the project.}%

\thanks{K. Min is with the Department of Brain and Cognitive Engineering, Korea University, Seoul 02841, South Korea. Email: {\tt\small ks\_min@korea.ac.kr}}%

\thanks{G.-H. Lee is with the Department of Computer and Radio Communications Engineering, Korea University, Seoul 02841, South Korea. Email: {\tt\small gunhlee@korea.ac.kr}}%

\thanks{S.-W. Lee is with the Department of Artificial Intelligence and the Department of Brain and Cognitive Engineering, Korea University, Seoul 02841, South Korea. Email: {\tt\small sw.lee@korea.ac.kr}}%
}

\maketitle
\thispagestyle{empty}
\pagestyle{empty}

\setlength{\tabcolsep}{8.5pt} 
\renewcommand{\arraystretch}{1.5} 

\begin{abstract}
Computer-aided diagnosis has recently received attention for its advantage of low cost and time efficiency. Although deep learning played a major role in the recent success of acne detection, there are still several challenges such as color shift by inconsistent illumination, variation in scales, and high density distribution. To address these problems, we propose an acne detection network which consists of three components, specifically: Composite Feature Refinement, Dynamic Context Enhancement, and Mask-Aware Multi-Attention. First, Composite Feature Refinement integrates semantic information and fine details to enrich feature representation, which mitigates the adverse impact of imbalanced illumination. Then, Dynamic Context Enhancement controls different receptive fields of multi-scale features for context enhancement to handle scale variation. Finally, Mask-Aware Multi-Attention detects densely arranged and small acne by suppressing uninformative regions and highlighting probable acne regions. Experiments are performed on acne image dataset ACNE04 and natural image dataset PASCAL VOC 2007. We demonstrate how our method achieves the state-of-the-art result on ACNE04 and competitive performance with previous state-of-the-art methods on the PASCAL VOC 2007.
\end{abstract}

\begin{figure*}[ht]
\centering
\scriptsize
\includegraphics[width=1.0\linewidth, height=6.9cm]{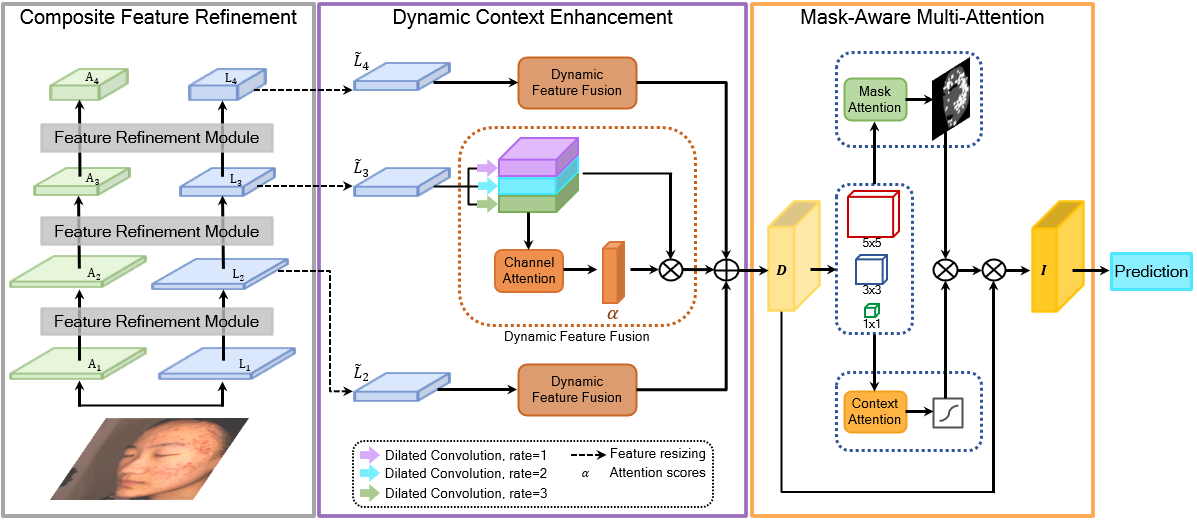} 
\caption{An overall architecture of ACNet. Our method consists of Composite Feature Refinement, Dynamic Context Enhancement, and Mask-Aware Multi-Attention. The feature map $I$ passed through the three preceding components are used for RPN\cite{ren2015faster} followed by classification and box regression.}
\label{fig1}
\end{figure*}

\section{\textbf{INTRODUCTION}}
Automated acne detection for computer-aided diagnosis is one of the important tasks in the medical field. Acne is a chronic skin disease which commonly exists on the face, neck, back, etc. Specific treatment is needed imminently for acne patients because the disease may lead to severe scars and pigmentation. Due to ambiguous appearance of acne, it is difficult for dermatologists to make a precise and time-efficient treatment decision. Besides, manual observation for acne is costly, since it relies on knowledge and experience of experts. Therefore, reliable computer-aided diagnosis for acne is necessary for effective and efficient treatment.

Recently, many studies have been conducted on acne analysis. However, there are still challenging problems of color shift, scale variation, and dense arrangement for acne detection. In real-world, exposed to uneven illumination, acne encounters low contrast or color transition.
They cause existing detectors to confuse the acne with other objects such as a mole on skin.
Also, the scale variation leads to performance degradation of detectors, especially missing small acne. 
Moreover, acne is often located in a group with high density, which causes mutual interference between features of each acne.

In the past years, most methods relied on hand-crafted features for acne detection. However, they showed unsatisfactory results due to the noise by color variation of the acne. To deal with the weakness, deep learning based methods are presented for detecting acne with the help of Convolution Neural Networks (CNNs). However, they perform classification on acne patch images or exploit generic object detectors with only hyper-parameter tuning without designing the architecture of detectors.

With this observation, we propose an acne detector called ACNet to detect acne under color shift, scale variation and dense arrangement. Our method consists of three components to address the above problems.
First, Composite Feature Refinement fuses deep semantics and shallow details to generate discriminative features. We devise combination of a composite backbone and a feature refinement module to deal with the problem of color transition.
Second, Dynamic Context Enhancement exploits dilated convolution with different dilation rates to aggregate multi-scale context features dynamically for alleviating the scale variation problem.
Third, Mask-Aware Multi-Attention suppresses noise and highlights object cues for densely distributed and small acne detection.

Qualitative and quantitative results show that our method outperforms previous state-of-the-art methods on ACNE04 dataset. 
Furthermore, ACNet exhibits comparable performance with existing generic object detectors on PASCAL VOC.
The main contributions of this paper are:
\begin{itemize}
\item We design Composite Feature Refinement to generate more powerful features with combination of a composite backbone and a feature refinement module which address the effect of uneven illumination.
\item To tackle the issue of scale variation, we introduce Dynamic Context Enhancement that dynamically controls receptive fields of multi-scale features.
\item For dense and small objects, we develop Mask-Aware Multi-Attention to diminish the negative impact of background noise and highlight foreground object cues.
\item We show that ACNet achieves the state-of-the-art performance on the ACNE04. Compared with the previous state-of-the-art detectors, favorable performance on PASCAL VOC validates the generality of our method.
\end{itemize}

\section{\textbf{Related Work}}
\subsection{Generic Object Detection}
Object detection is a fundamental research topic in computer vision with a wide range of applications\cite{kim2020few, zhang2020real, uehara2017object, lee1999integrated, lee1990translation}. Deep learning based object detectors are categorized into two-stage detectors and one-stage detectors. The two-stage detectors\cite{ren2015faster, NIPS2016_577ef115} firstly propose RoI (Region-of-Interest), and then the RoIs are further refined through a classifier and a box regressor. The one-stage detectors\cite{liu2016ssd, redmon2016you} directly perform classification and localization utilizing the predefined anchor boxes. Advantages of one-stage detectors are higher speed and less computation complexity than two-stage detectors. On the other hand, the two-stage detectors show better detection performance than that of the one-stage detectors. However, such detectors confront the aforementioned issues in terms of color, scale, and density, which limits adoption of generic object detectors for acne detection.

\subsection{Acne Detection}
Acne detection is divided into hand-crafted feature based approaches and deep learning based approaches.
For hand-crafted feature based methods, Chantharaphaichi \emph{et al}.\cite{chantharaphaichi2015automatic} utilizes the image processing technique based on HSV and gray scale color space. Then a binary thresholding is employed to find the acne spots.
Maroni \emph{et al}.\cite{maroni2017automated} uses a* channel of the CIE Lab color space to distinguish between healthy and unhealthy skin. Then the proposed system applies adaptive thresholding and performs the blob detection for marking acne spots based on Laplacian of Gaussian. However, these methods are still vulnerable to color variation by inconsistent lighting conditions.

For deep learning based methods, Chin \emph{et al}.\cite{chin2018facial} designed a facial pore detection method based on CNNs and investigated the performance of architectures with different depths.
Shen \emph{et al}.\cite{shen2018automatic} classifies the facial acne vulgaris with CNNs. The model crops the facial image into small patches and differentiates between skin and non-skin on the patch. In the end, the model predicts the acne type of the seven categories. Sophie Seite \emph{et al}.\cite{seite2019development} proposed an Artificial Intelligent Algorithm (AIA) to categorize the severity of facial acne into five classes utilizing Global Acne Severity Scale\cite{dreno2011development}. However, these methods are still not capable of detecting small and dense acne accurately.

\section{\textbf{Proposed Method}}
In this section, we introduce ACNet which consists of (1) Composite Feature Refinement, (2) Dynamic Context Enhancement, and (3) Mask-Aware Multi-Attention, as shown in Fig. \ref{fig1}. 

\subsection{Composite Feature Refinement}
\begin{figure}[t]
\centering
\includegraphics[width=1.0\linewidth]{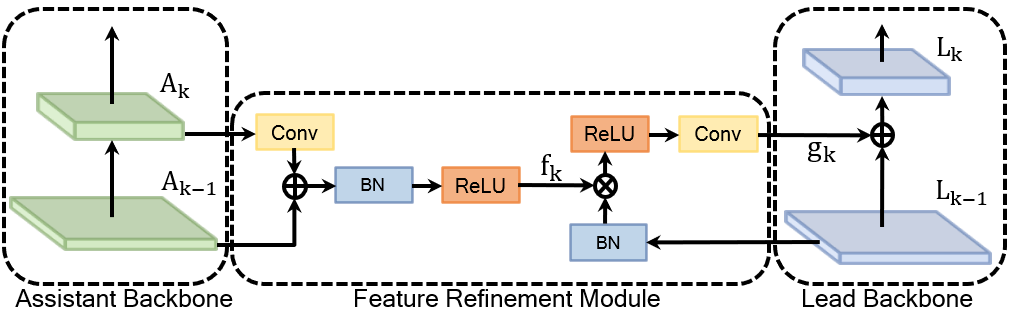}
\caption{Structure of Composite Feature Refinement (CFR). CFR combines three different features of Assistant Backbone and Lead Backbone. It injects then the fused features in the Lead Backbone.}
\label{fig2}
\end{figure}

As shown in Fig. \ref{fig2}, our Composite Feature Refinement comprises of a composite backbone and a feature refinement module. Taking inspiration from the structure of CBNet\cite{liu2020cbnet}, we exploit a dual backbone to extract discriminative features. Moreover, we design a feature refinement module to further enhance feature representation by fusing semantics and details for color shift by uneven illumination.

\subsubsection{Composite Backbone}
The composite backbone architecture consists of a lead backbone and an assistant backbone. It forms a composite connection between the parallel layers of two successive backbones.
Specifically, each current level feature of the assistant backbone is provided to the previous level of the lead backbone's layer iteratively. The output features of lead backbone are exploited for detection.

\subsubsection{Feature Refinement Module}
The dual backbone directly builds composite connections from deep layers to shallow layers to improve semantic information.
However, this fusion style interferes the detail information of shallow layers, which achieves the performance far from optimal.
Therefore, we design the feature refinement module to enhance the discrimination ability of the dual backbone by providing details of low-level features. Additionally, this module improves the small acne detection with the enrichment of detail features.

We employ upsampling and BatchNorm (BN) for each $k^{th}$ layer in assistant backbone, $A_{k}$. The generated layer and a previous layer $A_{k-1}$ of the assistant backbone are aggregated by element-wise summation followed by BN operation and ReLU activation. Fused features $f_{k}$ are defined as:

\begin{equation}
\begin{aligned}
    f_{k} = \gamma(\beta(A_{k-1} \oplus \phi_{k}(A_{k}))),\\
\end{aligned}
\end{equation}
where $\phi_{k}(.)$ denotes the operation including the serial bilinear interpolation and 1x1 $conv$ layer, $\beta$ is the BN, and $\gamma$ is the ReLU. To inject the fused features $f_{k}$ into the lead backbone, features $L_{k-1}$ from a $k$$-$$1^{th}$ shallow layer of the lead backbone is passed through BN. Afterwards, the modulated features and the fused features $f_{k}$ are combined using element-wise dot product. The final output $g_{k}$ of the feature refinement module is defined as:

\begin{equation}
\begin{aligned}
g_{k} = \varphi_{k}(\beta(L_{k-1}) \otimes f_{k}),
\end{aligned}
\end{equation}
where $\varphi_{k}(.)$ denotes the operation including the serial ReLU activation and 3$\times$3 $conv$ layer.

\subsection{Dynamic Context Enhancement}
Recent object detectors combine low-level details and high-level semantics to cope with scale variation. However, these detectors neglect that feature maps from each layer have different receptive fields. Therefore, we design Dynamic Context Enhancement to dynamically control the receptive fields of multi-scale feature maps by learning attention scores deploying a channel attention block. The pipeline consists of the two steps: feature resizing and dynamic feature fusion.

\subsubsection{Feature Resizing}
We use feature maps of different scales and denote each feature map at level $i$ ($i$ $\in$ {2,3,4}) as $L_{i}$. Because the features at each level have inconsistent channels and resolutions, we correspondingly employ the resizing strategy for $L_{2}$ and $L_{4}$ to merge with $L_{3}$.
$L_{2}$ is rescaled to the shapes of $L_{3}$ by applying an adaptive average pooling and a 1$\times$1 convolution. To obtain the same spatial dimensions of $L_{3}$, we apply bilinear interpolation and a 1$\times$1 convolution on $L_{4}$.

\begin{figure}[t]
\centering
\includegraphics[width=1.0\linewidth]{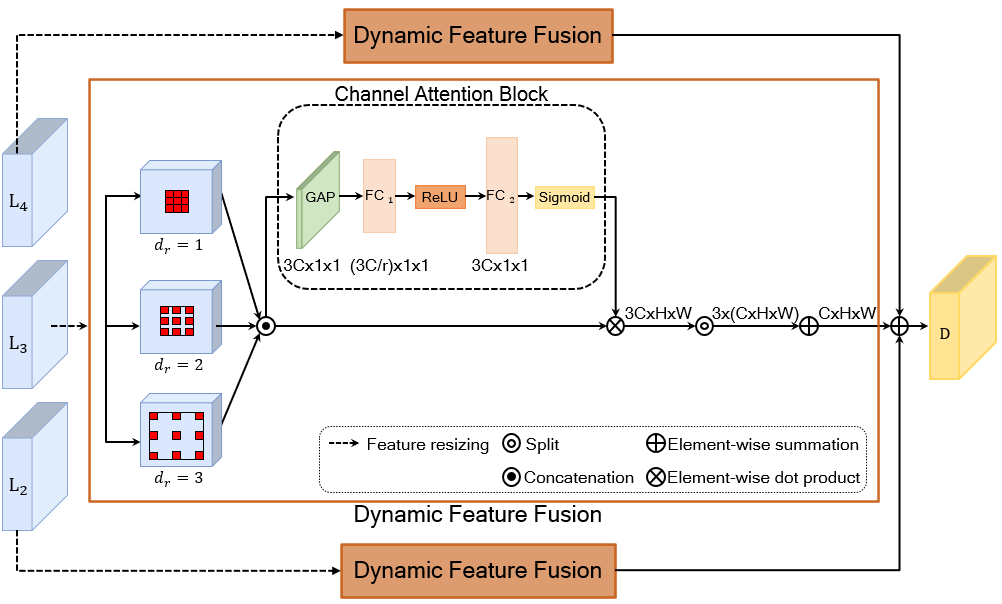}
\caption{Structure of Dynamic Context Enhancement (DCE). DCE leverages dilated convolution with different dilation rates to adjust receptive fields exploiting a channel attention block.}
\label{fig3}
\end{figure}

\subsubsection{Dynamic Feature Fusion}
Let $\widetilde{L}_{i}$ denote a feature map resized from $L_{i}$. We propose to dynamically adjust receptive fields of $\widetilde{L}_{i}$ as follows:

\begin{equation}
\begin{aligned}
D_{i} = \mathcal{H}(\widetilde{L}_{i}) = \psi(\mathcal{F}_{cat}^{i} \otimes \alpha^{i}),
\end{aligned}
\end{equation}

where $\mathcal{H}(\cdot)$ implies the Dynamic Feature Fusion operation. The output feature map $D_{i}$ of the operation is generated applying attention scores $\alpha^{i}$ on concatenated features 
$\mathcal{F}_{cat}^{i}$ which are combined using the output from different dilated convolutions. $\psi(.)$ denotes the operation including the serial split and element-wise sum operations. The concatenated features $\mathcal{F}_{cat}^{i}$ are defined as:

\begin{equation}
\begin{aligned}
    \mathcal{F}_{cat}^{i} = \lbrack\mathcal{C}_{d=1}(\widetilde{L}_{i}),\ \mathcal{C}_{d=2}(\widetilde{L}_{i}),\ \mathcal{C}_{d=3}(\widetilde{L}_{i})\rbrack.
\end{aligned}
\end{equation}

Here, [$\cdot$] denotes concatenation operation and $\mathcal{C}_{d=d_{r}}(\cdot)$ denotes dilated convolution operation with dilation rate $d_{r}$. The $\widetilde{L}_{i}$ undergoes dilated convolution with dilation rate 1, 2, and 3 to obtain different receptive fields, then followed by concatenation operation. Finally, attention scores $\alpha_{i}$ are applied to the concatenated features $\mathcal{F}_{cat}^{i}$ using element-wise dot product. The scores determine the importance of channels considering receptive fields in a channel attention block exploiting SENet\cite{hu2018squeeze} as follows:

\begin{equation}
\begin{aligned}
    \alpha^{i} = \sigma(W_{2}\cdot\delta(W_{1}\cdot(\mathcal{P}_{avg}(\mathcal{F}_{cat}^{i})))),
\end{aligned}
\end{equation} 

where $W_{1}\in \mathbb{R}^{C/r*C}$ denotes the parameters of the first fully connected layer $fc_{1}$, $W_{2} \in \mathbb{R}^{C*C/r}$ denotes the parameters of the second fully connected layer $fc_{2}$, $\mathcal{P}_{avg}(.)$ indicates average pooling, $\delta(.)$ is ReLU operation, and $\sigma(.)$ represents sigmoid. In the channel attention block, we set the reduction ratio to $r$=16.

With this method, the outputs $D_{2}$, $D_{3}$, and $D_{4}$ have enhanced receptive fields and are aggregated using element-wise summation operation. The detailed structure of Dynamic Context Enhancement is illustrated in Figure 3.

\begin{figure}[t]
\centering
\includegraphics[width=1.0\linewidth]{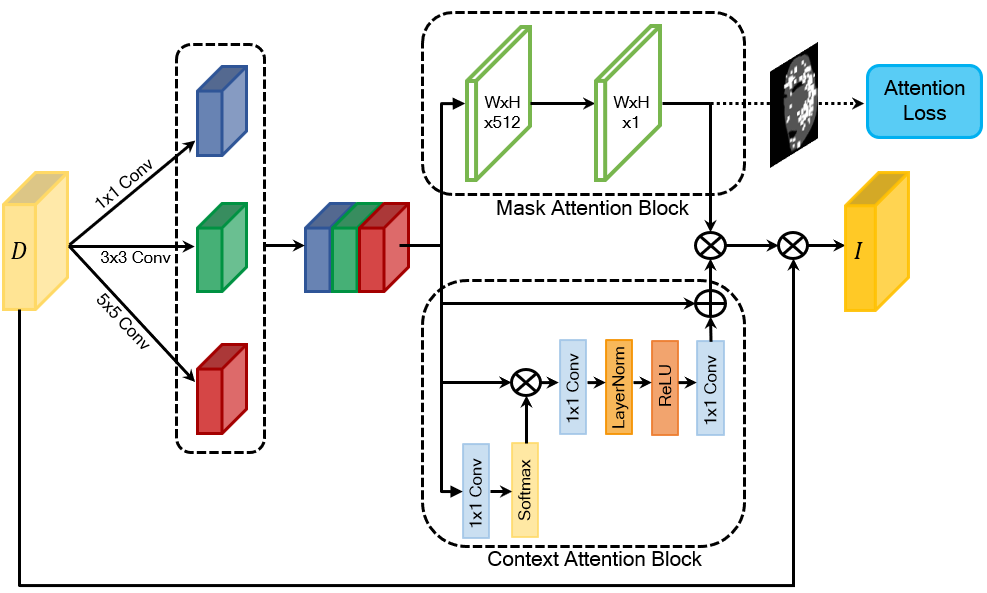}
\caption{Structure of Mask-Aware Multi-Attention (MAMA). MAMA consists of streamlined inception network, mask attention block, and context attention block.}
\label{fig4}
\end{figure}

\subsection{Mask-Aware Multi-Attention}
The Mask-Aware Multi-Attention detects objects of dense arrangement and small scale by emphasizing object-cues and reducing redundant noise. Although the Dynamic Context Enhancement generates features with increased receptive fields, excessive noise still remains in the features, which misleads to achieve sub-optimal performance. Thus, we employ a streamlined inception network\cite{szegedy2015going} to capture diverse shapes of acne using different ratio convolution kernels. Then, we exploit a mask attention block and a context attention block concurrently on output features of the streamlined inception network. The attention blocks increase the strong semantics and weaken the negative effect of noise.

\subsubsection{Mask Attention Block} 
Contrary to unsupervised attention structures\cite{hu2018squeeze ,woo2018cbam}, we design a mask attention block in a supervised manner. Redundant noise can disturb the object cues, which results in missing small acne on the face or detecting false positives. Therefore, it is essential to remove redundant information of non-acne area and enhance the salient features of the acne area. To focus on the acne cue on the face, we make use of the acne mask. In the mask attention block, the feature map $D$ passes through the inception network, and then a single channel saliency map is generated through consecutive convolution operations. Then, sigmoid operation is employed on the saliency map and the map undergoes element-wise dot product. In the end, the new feature map $I$ is created, as shown in Fig. \ref{fig4}. To guide the features to learn in a supervised fashion, we utilize a binary map as a label based on the ground truth, and then use binary cross-entropy loss between the saliency map and the binary map as attention loss.

\subsubsection{Context Attention Block}
Acne detection is challenging because of similar appearance with non-acne. On the face, the acne is often indistinguishable from the background including moles.
To address the problem, we exploit the Global Context block (GC block)\cite{cao2019gcnet} to amplify the contextual features relevant to the acne. To pay attention to sub-regions which have more relevance, we capture long-range dependencies among responses at all positions. It leads to enrichment of acne features in limited spatial resolution and distinction between acne and non-acne.

\subsection{Loss Function}
For detection, we employ a multi-task loss function which is defined as follows:
\begin{equation}
\begin{aligned}
    L = &\frac{1}{N_{cls}}\sum_{i}L_{cls}(p_{i}, p_{i}^{*}) \ +\\ &\frac{1}{N_{reg}}\sum_{i}p_{i}^{*}L_{reg}(t_{i}, t_{i}^{*}) \ +\\
    &\frac{1}{h*w}\sum_{i}^{h}\sum_{j}^{w}L_{att}(m_{ij}, m_{ij}^{*}),
\end{aligned}
\end{equation}
where $p_{i}$ is the predicted probability of classification and $p_{i}^{*}$ represents the ground-truth label which is 1 if the anchor is positive, 0 otherwise. $t_{i}$ is a vector representing the predicted offset vectors and $t_{i}^{*}$ is that of ground-truth. $m_{i}$, $m_{i}^{*}$ denote the mask's pixel of the ground-truth and the prediction.

The classification loss $L_{cls}$ is log loss for acne and background. $N_{cls}$ indicates the number of anchors. The regression loss $L_{reg}$ is smooth $L_{1}$ loss defined in Fast R-CNN\cite{girshick2015fast}. $N_{reg}$ denotes the number of anchors assigned to positive anchors. The attention loss $L_{att}$ is pixel-wise sigmoid cross entropy. $h$ and $w$ represent the height and width of the mask.

\section{\textbf{Experiments}}
\begin{figure*}[t]
\centering
\includegraphics[width=0.8\linewidth]{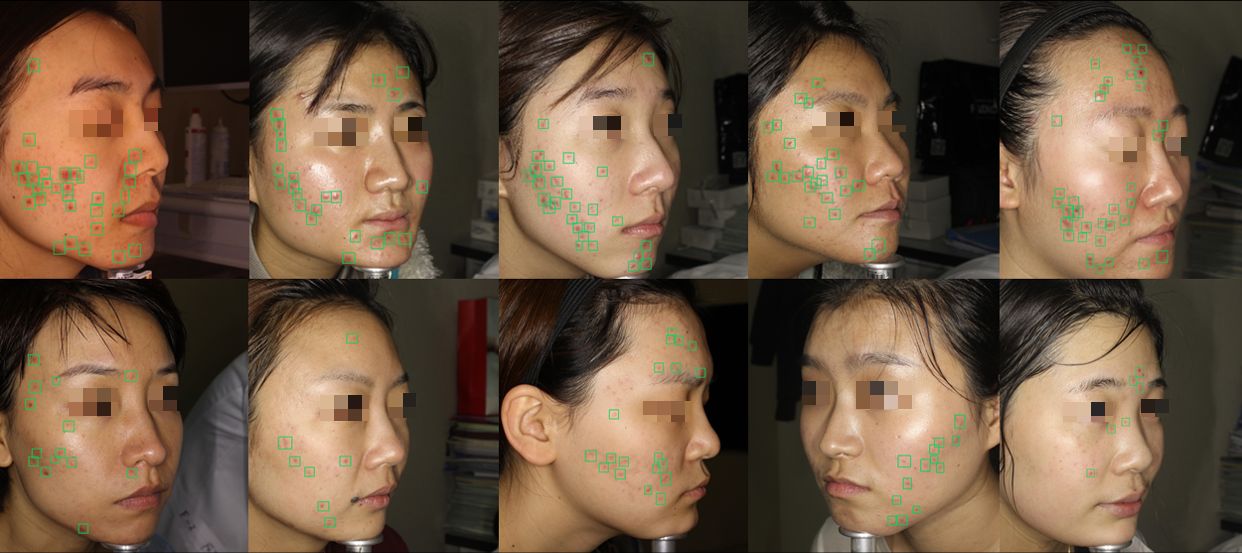}
\caption{Qualitative detection results of our method on the ACNE04 testset.}
\label{fig5}
\end{figure*}

The experiments are implemented by Pytorch on two Nvidia Titan V GPUs for training. 
We perform experiments on both ACNE04\cite{wu2019joint} and PASCAL VOC 2007\cite{everingham2010pascal}.
The evaluation metric adopted for performance is mean Average Precision (mAP).
Our framework is mainly aimed to detect facial acne, so the experimental result of the ACNE04 is used as the main evaluation criteria.
To verify the generality of our method, we also compare our detector with previous state-of-the-art methods on PASCAL VOC 2007.

\subsection{Experiments on Acne Images}
\subsubsection{Datasets and Protocols}
The ACNE04 is for facial acne detection. It contains 1,457 front and profile face images taken from the patients.
The image size varies from around 300$\times$300 to 4,000$\times$4,000 pixels and contains acne with a wide range of scales.
These images are annotated by professional dermatologists using a single object category.
The fully annotated ACNE04 contains 18,983 bounding boxes of acne.
The split ratios of the training and test images are 80\% and 20\% respectively.
We crop the images into 640$\times$640 subimages with an overlap of 320 pixels. No additional cropping is implemented on the image under the size of 640$\times$640.
\newline
\indent Our framework uses a dual ResNet-101 backbone pre-trained on ImageNet.
We perform the experiments on two datasets and use Stochastic Gradient Descent as an optimizer over 2 GPUs with a total of 4 images per minibatch (2 images per GPU). All models are trained for 30 epochs with an initial learning rate of 1e-4, which is then decreased by a factor of 10 after every 10 epochs. Only horizontal image flipping is performed for data augmentation. The anchor stride is set to 8 and the aspect ratios are set to [0.5, 1, 2]. We set the base anchor size to 256 and the anchor scales from 2$^{-3}$ to 2$^{0}$. To reduce the overlapped targets, we replace Non-Maximum Suppression (NMS) by Soft Non-Maximum Suppression (Soft-NMS)\cite{bodla2017soft}.

\renewcommand{\arraystretch}{1} 
\renewcommand{\tabcolsep}{2mm}
\begin{table}[t]
\caption{Acne detection performance (mAP) on ACNE04 testset. ResNet101$^{\star}$ represents our dual ResNet101 backbone.}
\begin{center}
{\small
\begin{tabular}{ll|c}
\Xhline{3\arrayrulewidth}
\multicolumn{1}{l}{Method} & \multicolumn{1}{l}{Backbone} & \multicolumn{1}{c}{mAP} \\ \Xhline{3\arrayrulewidth}
\multicolumn{1}{l}{Faster R-CNN\cite{ren2015faster}} & \multicolumn{1}{l}{ResNet101} & \multicolumn{1}{c}{10.3} \\
\multicolumn{1}{l}{R-FCN \cite{NIPS2016_577ef115}} & \multicolumn{1}{l}{ResNet101} & \multicolumn{1}{c}{14.0} \\
\multicolumn{1}{l}{Rashataprucksa $et \ al$. \cite{rashataprucksa2020acne}} & \multicolumn{1}{l}{ResNet101} & \multicolumn{1}{c}{14.7} \\
\multicolumn{1}{l}{ACNet} & \multicolumn{1}{l}{ResNet101$^{\star}$} & \multicolumn{1}{c}{\textbf{20.5}} \\
\Xhline{3\arrayrulewidth}
\end{tabular}
}
\label{tab1}
\end{center}
\end{table}

\begin{table}[t]
\caption{Ablation studies on ACNE04 testset. We use the Faster R-CNN [9] as the baseline, and gradually include Composite Backbone (CB), Feature Refinement Module (FRM), Dynamic Context Enhancement (DCE), and Mask-Aware Multi-
Attention (MAMA) for ablation studies.}

\renewcommand{\arraystretch}{1}
\renewcommand{\tabcolsep}{1.5mm}
\begin{center}
{\small
\begin{tabular}{c|cccc|c}
\Xhline{3\arrayrulewidth} 
\multicolumn{1}{l}{} & {CB} & {FRM} & {DCE} & {MAMA} & {mAP} \\ 
\Xhline{3\arrayrulewidth}
\multicolumn{1}{l}{Baseline} & {} & {} & {} & {} & {10.3}\\
\Xhline{1\arrayrulewidth}
\multicolumn{1}{c}{(a)} & {} & {\checkmark} & {\checkmark} & {\checkmark} & {15.1}\\
\multicolumn{1}{c}{(b)} & {\checkmark} & {} & {} & {} & {15.2}\\
\multicolumn{1}{c}{(c)} & {\checkmark} & {\checkmark} & {} & {} & {15.7}\\
\multicolumn{1}{c}{(d)} & {\checkmark} & {} & {\checkmark} & {} & {16.2}\\
\multicolumn{1}{c}{(e)} & {\checkmark} & {} & {} & {\checkmark} & {16.7}\\
\multicolumn{1}{c}{(f)} & {\checkmark} & {\checkmark} & {\checkmark} & {} & {17.9}\\
\multicolumn{1}{c}{(g)} & {\checkmark} & {\checkmark} & {} & {\checkmark} & {18.5}\\
\multicolumn{1}{c}{(h)} & {\checkmark} & {\checkmark} & {\checkmark} & {\checkmark} & {\textbf{20.5}}\\
\Xhline{3\arrayrulewidth}
\end{tabular}
}
\label{tab2}
\end{center}
\end{table}

\subsubsection{Ablation Study}
We use Faster-RCNN\cite{ren2015faster} with RoI align as the baseline. For fair comparison, ablation experiments are conducted with same hyper-parameter settings.

\textbf{Effect of Composite Feature Refinement.}
We first replace ResNet101 backbone with dual ResNet101 backbone in Composite Feature Refinement.
As shown in Table \ref{tab2}, the dual backbone improves the baseline method by 4.9 mAP (Table \ref{tab2}(b)). 
It benefits from that the connections between a lead backbone and an assistant backbone enrich the feature representation.
Feature Refinement Module further improves the performance from 15.2 to 15.7 (Table \ref{tab2}(c)). This slight gain in accuracy is achieved through utilization of both semantic and detail information from the assistant backbone.

\begin{table}[t]
\caption{Ablation studies of Dynamic Context Enhancement (DCE) on ACNE04 testset. Layers are aggregated by element-wise summation without DCE.}

\renewcommand{\arraystretch}{1} 
\renewcommand{\tabcolsep}{5mm}
\begin{center}
{\small
\begin{tabular}{c|c|c}
\Xhline{3\arrayrulewidth} 
{Layers} & {Method} & {mAP} \\ 
\Xhline{3\arrayrulewidth}
\multirow{2}{*}{$L_{1}$ - $L_{4}$} & DCE (w/o) & 18.1 \\ & DCE (w) & 20.0\\
\Xhline{1\arrayrulewidth}
\multirow{2}{*}{$L_{2}$ - $L_{4}$} & DCE (w/o) & 18.5 \\ & DCE (w) & \textbf{20.5}\\
\Xhline{1\arrayrulewidth}
\multirow{2}{*}{$L_{3}$ - $L_{4}$} & DCE (w/o) & 18.2 \\ & DCE (w) & 20.2\\
\Xhline{3\arrayrulewidth}
\end{tabular}
}
\label{tab3}
\end{center}
\end{table}

\textbf{Effect of Dynamic Context Enhancement.}
We propose Dynamic Context Enhancement to control the receptive field of each feature map exploiting 3$\times$3 convolution kernels with different dilation rate from 1 to 3. As shown in (Table \ref{tab2}(d) and (f)), using Dynamic Context Enhancement provides 1.0 and 2.2 mAP gains because of the adjusted receptive fields of objects in different scales. These results demonstrate the effectiveness of Dynamic Context Enhancement.

\textbf{Effect of Mask-Aware Multi-Attention.}
As discussed in Sec. \rom{3}. C, Mask-Aware Multi-Attention is beneficial to reduce the impact of noise and highlight the probable object regions. The method achieves 16.7 and 18.5 mAP on the detection performance with a large margin of 1.5 and 2.8 mAP (Table \ref{tab2}(e) and (g)). In Fig. \ref{fig5}, we can find that the ACNet detects more acne with small scale and in dense arrangement. We infer that mask attention weakens the background noise and context attention strengthens the acne features with high relevance. This combination of mask attention and context attention generates refined features, which improves the detection performance.

\renewcommand{\arraystretch}{1} 
\renewcommand{\tabcolsep}{3mm}
\begin{table}[t]
\caption{Performance on PASCAL VOC 2007 testset. ResNet101$^{*}$ represents our dual ResNet101 backbone.}
\begin{center}
\footnotesize{
\begin{tabular}{l@{/,}cc|c}
\Xhline{3\arrayrulewidth} 
\multicolumn{1}{l}{Method} & Backbone & Input Size & mAP \\ \Xhline{3\arrayrulewidth}
\multicolumn{1}{l}{\textbf{Single-Stage Detectors:}} & {} & {} & {} \\
\multicolumn{1}{l}{SSD300 [12]} & VGG16 & 300$\times$300 & 77.2 \\
\multicolumn{1}{l}{DSSD321 [23]} & ResNet101 & 321$\times$321 & 78.6 \\
\multicolumn{1}{l}{DES300 [24]} & ResNet101 & 300$\times$300 & 79.7 \\
\multicolumn{1}{l}{RefineDet300 [25]} & VGG16 & 300$\times$300 & 80.0 \\
\multicolumn{1}{l}{RFBNet300 [26]} & VGG16 & 300$\times$300 & 80.5 \\
\Xhline{3\arrayrulewidth}
\multicolumn{1}{l}{SSD512 [12]} & VGG16 & 512$\times$512 & 79.5 \\
\multicolumn{1}{l}{DSSD513 [23]} & ResNet101 & 513$\times$513 & 81.5 \\
\multicolumn{1}{l}{DES512 [24]} & VGG16 & 512$\times$512 & 81.7 \\
\multicolumn{1}{l}{RefineDet512 [25]} & VGG16 & 512$\times$512 & 81.8 \\
\multicolumn{1}{l}{RFBNet512 [26]} & VGG16 & 512$\times$512 & 82.1 \\
\Xhline{3\arrayrulewidth}
\multicolumn{1}{l}{\textbf{Two-Stage Detectors:}} & {} & {} & {} \\
\multicolumn{1}{l}{Faster R-CNN [9]} & ResNet101 & 1000$\times$600 & 76.4 \\
\multicolumn{1}{l}{R-FCN [11]} & ResNet101 & 1000$\times$600 & 80.5 \\
\multicolumn{1}{l}{ACNet} & ResNet101$^{\star}$ & 1000$\times$600 & 81.8\\
\multicolumn{1}{l}{CoupleNet [27]} & ResNet101 & 1000$\times$600 & 82.7 \\
\Xhline{3\arrayrulewidth}
\end{tabular}
}
\label{tab4}
\end{center}
\end{table}

\subsection{Experiments on Natural Images}
To demonstrate the generality of the proposed method, we perform experiments on the PASCAL VOC 2007 dataset as it contains small generic objects. As shown in Table \ref{tab4}, our ACNet achieves the test result of 81.8 mAP on PASCAL VOC 2007, surpassing most of the one-stage detectors, and even leading the majority of two-stage detectors. Our method has shown comparable performance with the previous state-of-the-art generic object detectors on PASCAL VOC 2007. We infer that it is because the PASCAL VOC 2007 scarcely has color shift and dense arrangement problems.

\section{\textbf{CONCLUSION}}
In this paper, we present the acne detector called ACNet designed to deal with imbalanced illumination, scale variation and dense arrangement. 
We introduce Composite Feature Refinement to produce discriminative features aggregating high-level semantics and low-level details.
Further, we propose Dynamic Context Enhancement, which is used to dynamically adjust the different receptive fields for addressing scale variation.
Finally, we design Mask-Aware Multi-Attention to reduce uninformative noise and highlight probable object regions to detect densely arranged and small objects.
Experiments show that our method achieves the state-of-the-art performance on the ACNE04 dataset.

\addtolength{\textheight}{-12cm}  

\bibliographystyle{IEEEtran.bst}
\bibliography{root.bbl}

\end{document}